\documentclass[10pt,twocolumn,letterpaper]{article}

\usepackage{iccv}
\usepackage{times}
\usepackage{epsfig}
\usepackage{graphicx}
\usepackage{amsmath}
\usepackage{amssymb}
\usepackage{bm}
\usepackage[ruled,linesnumbered]{algorithm2e}
\usepackage{multirow}

\usepackage[pagebackref=true,breaklinks=true,letterpaper=true,colorlinks,bookmarks=false]{hyperref}
\newcommand{\tabincell}[2]{\begin{tabular}{@{}#1@{}}#2\end{tabular}}
\newcommand\blfootnote[1]{%
  \begingroup
  \renewcommand\thefootnote{}\footnote{#1}%
  \addtocounter{footnote}{-1}%
  \endgroup
}
\iccvfinalcopy 

\def\iccvPaperID{2530} 

\ificcvfinal\fi

\begin{document}

\title{3DHacker: Spectrum-based Decision Boundary Generation  \\ for Hard-label 3D Point Cloud Attack}

\author{
Yunbo Tao\textsuperscript{1*} \ \
Daizong Liu\textsuperscript{2*} \ \
Pan Zhou\textsuperscript{1} \ \
Yulai Xie\textsuperscript{1$\dagger$} \ \
Wei Du\textsuperscript{1} \ \
Wei Hu\textsuperscript{2$\dagger$} \\
\textsuperscript{1}Hubei Key Laboratory of Distributed System Security, \\ Hubei Engineering Research Center on Big Data Security, \\ School of Cyber Science and Engineering, Huazhong University of Science and Technology \\
\textsuperscript{2}Wangxuan Institute of Computer Technology, Peking University
\\
{\tt\small \{tyb666, panzhou, ylxie, weidu666\}@hust.edu.cn  \quad dzliu@stu.pku.edu.cn \quad forhuwei@pku.edu.cn} 
}

\maketitle
\ificcvfinal\thispagestyle{empty}\fi

\begin{abstract}
\vspace{-10pt}
With the maturity of depth sensors, the vulnerability of 3D point cloud models has received increasing attention in various applications such as autonomous driving and robot navigation.
Previous 3D adversarial attackers either follow the white-box setting to iteratively update the coordinate perturbations based on gradients, or utilize the output model logits to estimate noisy gradients in the black-box setting.
However, these attack methods are hard to be deployed in real-world scenarios since realistic 3D applications will not share any model details to users.
Therefore, we explore a more challenging yet practical 3D attack setting, \textit{i.e.}, attacking point clouds with black-box hard labels, in which the attacker can only have access to
the prediction label of the input.
To tackle this setting, we propose a novel 3D attack method, termed \textbf{3D} \textbf{H}ard-label att\textbf{acker} (\textbf{3DHacker}), based on the developed decision boundary algorithm to generate adversarial samples solely with the knowledge of class labels.
Specifically, to construct the class-aware model decision boundary, 3DHacker first randomly fuses two point clouds of different classes in the spectral domain to craft their intermediate sample with high imperceptibility, then projects it onto the decision boundary via binary search. To restrict the final perturbation size, 3DHacker further introduces an iterative optimization strategy to move the intermediate sample along the decision boundary for generating adversarial point clouds with smallest trivial perturbations.
Extensive evaluations show that, even in the challenging hard-label setting, 3DHacker still competitively outperforms existing 3D attacks regarding the attack performance as well as adversary quality.
\end{abstract}

\vspace{-30pt}

\blfootnote{
\textsuperscript{$*$}Equal contributions. ~~~~\textsuperscript{$\dagger$}Corresponding authors.}
\section{Introduction}

Deep Neural Networks (DNNs) are known to be vulnerable to adversarial examples \cite{szegedy2013intriguing,goodfellow2014explaining}, which are indistinguishable from legitimate ones by adding trivial perturbations but often lead to incorrect model prediction.
Many efforts have been made into attacks on the 2D image field \cite{dong2018boosting,madry2017towards,kurakin2016adversarial,tu2019autozoom}. 
However, in addition to image-based 2D attacks, 3D point cloud attacks are also crucial in various safety-critical
applications such as autonomous driving \cite{chen2017multi,yue2018lidar,hu2023density}, robotic grasping \cite{varley2017shape,zhong2020reliable}, and face challenges in realistic scenarios. 

\begin{figure}[!t]
	\centering
	\includegraphics[width=0.47\textwidth]{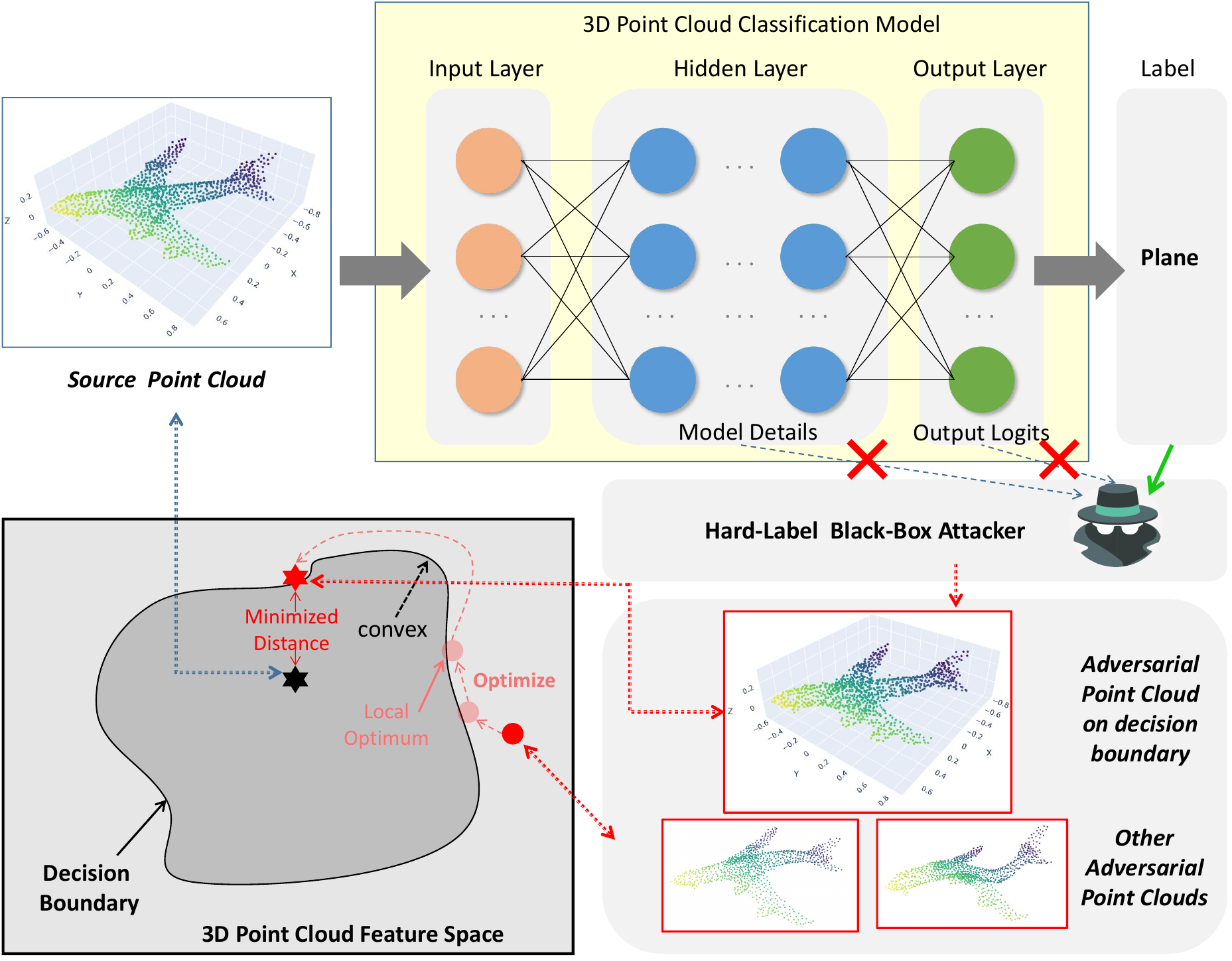}
	\caption{Illustration of our motivation. Our 3D hard-label setting cannot access any model details of the hidden layer or the logits of the output layer. To tackle this setting, we develop a spectrum-based decision boundary algorithm to iteratively optimize the best adversarial sample only with the knowledge of the predicted label.}
    \vspace{-10pt}
	\label{fig_1}
\end{figure}

Existing 3D point cloud attack methods \cite{tsai2020robust,zhao2020isometry,zhou2020lg,wen2020geometry}
generally delve into achieving high attack success rate and maintaining the imperceptibility of adversarial examples.
However, these 3D attack methods still suffer from two main limitations:
(1) Most of them are deployed in white-box setting where the attackers have the full knowledge of victim models including  network structure and weights. This setting makes the attacks less practical since most real-world 3D applications will not share their model details to users. 
(2) The quality of their adversarial examples are limited. Since most previous works simply utilize geometric distance losses or additional shape knowledge to implicitly constrain the perturbations according to gradient search or gradient optimization,
their adversarial examples fail to properly keep the original 3D object shape and easily have irregular surface or outliers. Although some attackers \cite{zhang2019adversarial,xiang2019generating,wen2020geometry,huang2022shape} try to modify few points or design geometry-aware perturbations, their adversarial samples are hard to achieve optimal due to the greedy search in the optimization process.

Based on the above observations, in this paper, we make the attempt to explore a more challenging yet practical 3D attack setting, \textit{i.e.}, attacking 3D point clouds with black-box hard labels, in which attackers can only have access to the final classification results of the model instead of the model details. 
Compared to previous general 3D black-box setting \cite{huang2022shape} that still has the knowledge of predicted logits scores of the input, our hard-label setting
is more difficult since we only access the prediction labels of the input and cannot rely on the changes of the final predicted logits for updating the perturbations.
To tackle this new attack setting while improving the quality of adversarial examples compared to previous works, we aim to exploit the decision boundary mechanism \cite{li2020qeba,li2022decision,li2021nonlinear} as the core idea to generate adversarial point clouds 
by determining the class-aware model decision boundary for searching the mispredicted boundary cloud with small distance to the source cloud, 
as shown in Figure~\ref{fig_1}. Based on the concave-convex decision boundary, it can easily optimize samples with lowest perturbations.

However, directly adapting previous 2D decision boundary mechanisms \cite{srinivasan2019black,ilyas2018black,li2022decision,li2020qeba,li2021nonlinear} into the 3D point cloud field may face several challenges:
(1) Previous 2D decision-boundary attackers \cite{li2020qeba,li2021nonlinear} generally generate adversarial images on decision boundary by calculating the weighted average of each pixel value between source and target images. However, 3D point cloud data mainly contain coordinates with optional point-wise values, which is unordered and irregularly sampled. 
Therefore, directly calculating the weighted average of the point coordinates between two point clouds would make no sense, since it will disarrange the relations of neighboring points and deform the 3D object shape.
Although one solution would be to solely add coordinate-wise perturbations to generate the decision boundary, this operation is not geometry-aware and easily leads to outliers and uneven point distribution (validated in our Experiment Section).
(2) In the 2D field, in addition to achieving high imperceptibility, the averaged image also preserves the semantic features of the original image since their pixel relations are implicitly maintained.
However, the structural representations of 3D point clouds in the latent space will be easily broken when modifying points in the data domain.
Therefore, it is critically challenging to maintain both smooth appearance and geometric characteristics for generating high-quality 3D adversarial examples.
(3) Almost all 2D decision boundary mechanisms solely utilize pixel-wise iterative walking to move the image along the decision boundary for optimization. However, directly applying the point-wise walking on 3D point clouds may stuck into the local optimum, since the optimized cloud on the decision boundary may not have the smallest perturbation and fail to overcome the large convex area without additional guidance as shown in Figure~\ref{fig_1}.

 To address the above challenges, we propose a novel spectrum-based decision boundary attack method, called \textbf{3D} \textbf{H}ard-label att\textbf{acker} (\textbf{3DHacker}).
 Instead of fusing the point cloud via coordinate-wise average operation, we propose to fuse point clouds in the spectral domain, which not only preserves the geometric characteristics of both point clouds in coordinate-aware data domain, but also produces the smooth structure of the generated sample for achieving high imperceptibility.
 Specifically, our 3DHacker consists of two parts-——\emph{boundary-cloud generation} with spectrum-fusion and \emph{boundary-cloud optimization} along the decision boundary. 
 (1) During the \emph{boundary-cloud generation},
 we aim to attack a benign source-cloud into the class of target-clouds at their decision boundary with trivial perturbations.
 To be specific, we first perform Graph Fourier Transform (GFT) on each point cloud to obtain its graph spectrum, then fuse the spectrum of source-cloud and target-clouds with proper fusion rate with further inverse GFT to generate corresponding candidate point clouds.
 In this manner, we can obtain the decision boundary between the point clouds of different class labels.
We select the best candidate with slightest distortion and project it on the decision boundary to obtain the boundary-cloud.
 (2) In \emph{boundary-cloud optimization} stage, we perform iterative walking strategy on the boundary-cloud along the decision boundary aiming at further minimizing the distance between boundary-cloud and source-cloud. Each iteration starts from a perturbation generated by gradient estimation and then reduces the distortion through binary searching coordinates in the adversarial region. To jump out of the local optimum during optimization, we further design a spectrum-wise walking strategy in addition to the point-wise one.
 The boundary cloud optimized by the above two walking strategies will be taken as the final adversarial sample of the source cloud with high imperceptibility.

To sum up, our main contributions are three-fold:
\vspace{-4pt}
\begin{itemize}
    \item We achieve 3D point cloud attack in a more challenging yet practical black-box hard-label setting, and introduce 
    a novel method 3DHacker based on decision boundary algorithm with point cloud spectrum.
\vspace{-4pt}
    \item We propose spectrum-based decision boundary generation for preserving high-quality point cloud samples. Moreover, we introduce a new generation pipeline for boundary point clouds, and propose a joint coordinate- and spectrum-aware iterative walking strategy to alleviate the local optimum problem.
\vspace{-4pt}
    \item Experimental results show that the proposed 3DHacker achieves 100\% of attack success rates with the competitive perturbations compared to existing attack methods, even in the more challenging hard-label setting.
\end{itemize}


\section{Related Work}

\noindent \textbf{Adversarial attack on 3D point cloud.}
Following previous studies on the 2D image field, many works \cite{xiang2019generating,wicker2019robustness,zhang2019adversarial,zheng2019pointcloud,tsai2020robust,zhao2020isometry,zhou2020lg} adapt adversarial attacks into the 3D vision community\cite{guo2021pct,zhao2021point,yu2022point,zhang2022pvt,wang2019dynamic,uy2019revisiting,chang2015shapenet,qi2017pointnet,qi2017pointnet++,wu20153d}. 
\cite{xiang2019generating} propose point generation attacks by adding a limited number of synthesized points/clusters/objects to a point cloud.
\cite{zhang2019adversarial} utilize gradient-guided attack methods to explore point modification, addition, and deletion attacks.
Their goal is to add or delete key points, which can be identified by calculating the label-dependent importance score referring to the calculated gradient.
Recently, more works \cite{liu2019extending,zhang2019defense,tsai2020robust} adopt point-wise perturbation by changing their xyz coordinates, which are more effective and efficient. 
\cite{liu2019extending} modify the FGSM strategy to iteratively search the desired pixel-wise perturbation.
\cite{tsai2020robust} adapt the C\&W strategy to generate adversarial examples on point clouds and proposed a perturbation-constrained regularization in the overall loss function. 
Besides, some works \cite{lee2020shapeadv,kim2021minimal,shi2022shape,dong2022isometric} attack point clouds in the feature space and target at perturbing less points with lighter distortions for an imperceptible adversarial attack. However, the generated adversarial point clouds of all above methods often result in messy distribution or outliers, which is easily perceivable by humans and defended by previous adversarial defense methods \cite{zhou2019dup,perez20223deformrs,li2022robust}.
Although \cite{wen2020geometry} improves the imperceptibility of adversarial attacks via a geometry-aware constraint, the adversarial samples sometimes also deform local surfaces and are still noticeable. 

\noindent \textbf{Decision boundary based 2D attacks.}
Decision boundary based attack method \cite{brendel2017decision} is widely used in 2D filed, which is an efficient framework that uses final decision results of a classification model to implement hard-label black-box attack. In 2D field, the decision boundary attack process starts with two origin images called 
\emph{source-image} and \emph{target-image} with different labels. Next, it performs a binary search to obtain a \emph{boundary-image} on the decision boundary between \emph{source-image} and \emph{target-image}. Then, the random walking algorithm is conducted on \emph{boundary-image} with the goal of minimizing the distance towards \emph{target-image} and maintaining the label same as \emph{source-image}. 
Based on this general attack framework, various 2D decision boundary based attacks are proposed committed to improve the random walking performance and query efficiency.
\cite{brunner2019guessing,srinivasan2019black} propose to
choose more efficient random perturbation including Perlin noise and DCT in random walking step instead of Gaussian perturbation. \cite{chen2020hopskipjumpattack} conduct a gradient estimation method using Monte-carlo sampling strategy instead of random perturbation. \cite{li2020qeba,li2021nonlinear,li2022decision} improve the gradient estimation strategy through sampling from representative low-dimensional subspace. 
However, to our best knowledge, there is no decision boundary based attack in 3D vision community so far, and directly adapting these 2D methods into 3D field may face many challenges. 

\begin{figure*}[ht]
	\centering
 \vspace{-4pt}
	\includegraphics[width=\textwidth]{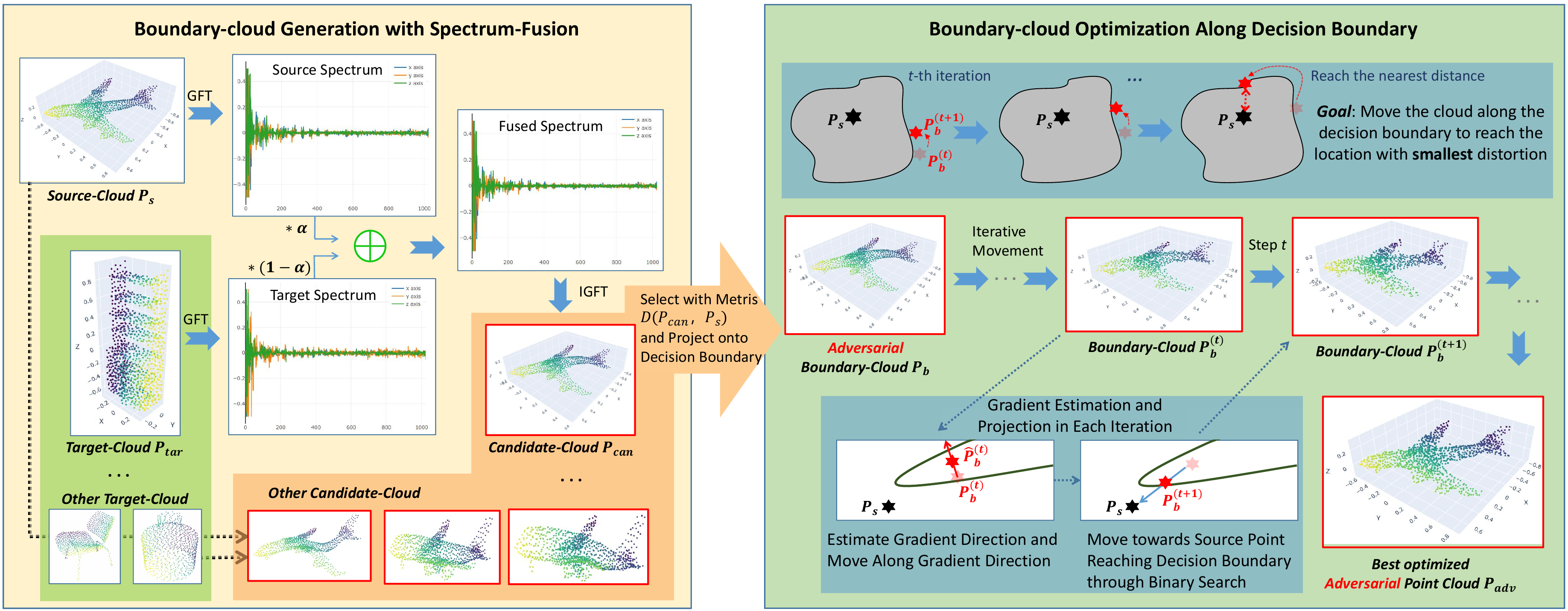}
	\caption{Overall pipeline of our proposed 3DHacker. The boundary-cloud generation module first fuses the source cloud with a set of target clouds in the spectral domain to construct candidate clouds, then selects the best one and projects it onto the decision boundary to obtain the boundary cloud via binary search. After that, the boundary-cloud optimization module iteratively moves the boundary cloud along the decision boundary via a joint coordinate-spectrum iterative walking strategy to achieve the best place with smallest distortion.}
 \vspace{-12pt}
	\label{fig_2}
\end{figure*}

\section{Methodology}
\subsection{Notations and Problem Definition}
\noindent \textbf{3D point cloud task.}
Generally, a point cloud consists of an unordered set of points $\bm{P}=\{\bm{p}_i\}_{i=1}^n \in \mathbb{R}^{n \times 3}$ sampled from the surface of a 3D object or scene, where each point $\bm{p}_i \in \mathbb{R}^{3}$ is a vector that contains the coordinates $(x,y,z)$ of point $i$, and $n$ is the number of points. 
In this paper, we mainly focus on the basic point cloud classification task. Given a point cloud $\bm{P}$ as input, a learned classifier $f(\cdot)$ predicts a vector of confidence scores $f(\bm{P})\in {\mathbb{R}^{C}}$. The final predicted label is $y=F(\bm{P})=argmax_{i\in[C]}f(\bm{P})_i\in { Y}, {Y}= \{1,2,3,...,C\}$ that represents the class of the original 3D object underlying the point cloud, where $C$ is the number of classes.

\noindent \textbf{Hard-label attack setting on 3D point cloud.} 
In the hard-label attack setting, the attackers can only access the final predicted label $y=F(\bm{P})$ instead of the model parameters $f(\cdot)$ and the confidence scores (logits) $f(\bm{P})$ for gradient optimization. 
Therefore, we aim to search the decision boundary of the classification model to determine the difference between 3D object classes, and utilize the decision boundary to estimate the positive gradient for updating the perturbation.
The perturbed adversarial sample $\bm{P}_{adv}$ is slightly generated from correctly classified source sample $\bm{P}_s$ labeled as $y_{true}$ such that the predicted label $y_{adv}=F(\bm{P}_{adv})\ne y_{true}$. We define an indicator function $\varphi(\cdot)$ of a successful attack as:
\begin{equation}\label{eq1}
\varphi(\bm{P}_{adv})\equiv\begin{cases}
1,  & \text{if } y_{adv} \ne y_{true},  \\
-1, & \text{if } y_{adv} = y_{true},
\end{cases}
\end{equation}
where $y_s$ is the true label of source point cloud $\bm{P}_s$, $y_{adv} = F(\bm{P}_{adv})$ is the label of $\bm{P}_{adv}$ predicted by the 3D model.

\subsection{Overview of Our 3DHacker Framework}
\label{section3.2}
To tackle the challenging 3D hard-label attack setting, in this section, we introduce the overall pipeline of our proposed 3DHacker, which starts with one source-cloud $\bm{P}_s$ of benign label $y_{true}$ that is correctly classed by model $f(\cdot)$, and a group of target-clouds $\{\bm{P}_{tar}\}$ of class labels $\{y_{tar}\}$, $y_{tar} \ne y_{true}$. 3DHacker aims to generate a boundary cloud on the model decision boundary between the source and target point clouds as the adversarial sample, which has a different class label of the source cloud while sharing the same object shape with it.
As shown in Figure~\ref{fig_2}, our 3DHacker
consists of two main parts —— A \textbf{boundary-cloud generation} module is first utilized to produce a high-quality adversarial point cloud on the decision boundary, then a \textbf{boundary-cloud optimization} module is exploited to further optimize the adversarial point cloud along the decision boundary for achieving smallest perturbations. 

\noindent  \textbf{First step: Boundary-cloud generation with spectrum-fusion.} 
In this step, we first perform the fusion method between a source cloud $\bm{P}_s$ and multiple target clouds $\bm{P}_{tar}$ to obtain candidate point clouds $\bm{P}_{can}$ to construct the decision boundary. Then we 
select a best adversarial $\bm{P}_{can}$ such that $\varphi(\bm{P}_{can})=1$ and $D(\bm{P}_s,\bm{P}_{can})$ is the smallest, where $D$ is the distance metric. After that, we project the best $\bm{P}_{can}$ onto the decision boundary via binary search algorithm to obtain the boundary cloud $\bm{P}_{b}$.

Previous 2D decision boundary mechanisms generally fuse the source and target samples via the pixel-wise average operation.
However, directly applying this strategy into 3D domain is ineffective since 3D coordinate-wise fusion between point clouds will result in outlier problem and destroy the 3D object geometric shape, leading to poor imperceptibility.
To address this issue, we introduce a new spectrum-fusion method, which fuses point clouds in the spectral domain to preserve the geometric topology. Instead of fusing point-wise coordinations, the spectrum-fusion method first transforms both source and target point clouds into the spectral domain for representing their geometric characteristics, then fuses their spectral contexts and transforms the fused geometric characteristics back to the data domain as the adversarial sample $\bm{P}_{can}$. In this manner, the fused point cloud not only preserves the specific characteristics of the original point clouds, but also has more smooth geometric surface due to the spectrum advantages \cite{hu2021graph,hu2022exploring}, leading to high imperceptibility.
Although with a slight and smooth perturbation, $\bm{P}_{can}$ still possesses a strong potential to confuse the model $f(\cdot)$ with a structure distortion learned from spectral characteristics of the $\bm{P}_{tar}$. 

\noindent  \textbf{Second step: Boundary-cloud optimization along the decision boundary.} 
Although we can obtain the boundary cloud $\bm{P}_{b}$ in the first step, this $\bm{P}_{b}$ is not the best optimal one due to the specific concave-convex structure of the decision boundary.
Therefore, we propose to update $\bm{P}_{b}$ along the decision boundary via the iterative walking strategy to search for a high-quality $\bm{P}_{b}$ that has the lowest geometric distance to the source cloud.

Previous 2D decision boundary mechanisms
generally conduct the iterative walking strategy with pixel-wise offsets, which can not be directly applied into the 3D domain since point-wise coordinate walking will lead to local optimum object shape.
The reason is,
although this boundary cloud is well optimized and still remains adversarial, 
it may get stuck in the local concave area with a large neighboring convex area, and the estimated gradients towards this convex will lead to a larger geometric distance to the source cloud.
To alleviate this issue, we extend the point-wise coordinate walking with a new spectrum walking as additional guidance for jumping out of such local optimum. 
That is, we update the boundary cloud $\bm{P}_{b}$ along the decision boundary with coordinate modification for shape refinement and spectrum modification for maintaining the geometric smoothness and high imperceptibility.
At last, the updated $\bm{P}_{b}$ on the decision boundary is our final generated adversarial sample.
In the following, we will provide more details about each step.

\subsection{Boundary-Cloud Generation with Spectrum-Fusion}
\label{section:3.3}

Given the source cloud $\bm{P}_s$ and $T$ number of target clouds ${\bm{P}_{tar}}$, we aim to generate a desired boundary-cloud $\bm{P}_b$ of label $y_b\ne y_{true}$ with a minimized distance $D(\bm{P}_s,\bm{P}_b)$.
Specifically, for each pair of $\bm{P}_s$ and ${\bm{P}_{tar}}$, we first conduct Graph Fourier Transform (GFT) \cite{hu2021graph} to obtain their corresponding spectrum coefficient vector $\hat{\bm{x}}_s$ and ${\hat{\bm{x}}_{tar}}$. Considering that low-frequency components mainly represent the basic 3D object shape while high-frequency components encode find-grained details \cite{hu2022exploring}, we then fuse their spectrum coefficient vectors with different fusion weights for low and high frequencies, and perform inverse GFT (IGFT) to transform the fused spectrum coefficient vector back to data domain for obtaining the candidate-cloud $\bm{P}_{can}$. In this way, we can get $T$ candidate-clouds $\bm{P}_{can}$ for $T$ target clouds ${\bm{P}_{tar}}$. The above process can be formulated as:
\begin{equation}\label{eq2}
\begin{aligned}
\hat{\bm{x}}_{low} &=\alpha_{low} \text{GFT}(\bm{P}_s)_L+(1-\alpha_{low})\text{GFT}(\bm{P}_{tar})_L, \\
\hat{\bm{x}}_{high}&=\alpha_{high} \text{GFT}(\bm{P}_s)_H+(1-\alpha_{high})\text{GFT}(\bm{P}_{tar})_H, \\
\bm{P}_{can}&=\text{IGFT} (\hat{\bm{x}}_{low}\boxplus\hat{\bm{x}}_{high}),
\end{aligned}
\end{equation}
where $GFT(\bm{P}_s)_L$,$GFT(\bm{P}_s)_H$ denotes conducting GFT on $\bm{P}_s$ and splitting its spectrum coefficient into the low $L$ and high $H$ frequencies. $\hat{\bm{x}}_{low}\boxplus\hat{\bm{x}}_{high}$ denotes piecing two coefficient vectors into a complete one. For spectrum-fusion, we use different fusion weights $\alpha_{low}$ and $\alpha_{high}$ for low and high spectrum frequencies since the lowest 32 frequencies account for almost 90\% of energy and the other 992 spectrum frequencies account for 10\% of energy \cite{liu2022point}, which lead to point clouds more sensitive to modifications on the lowest 32 frequencies than higher frequencies.

After obtaining $T$ candidate clouds, we conduct 
adversarial evaluation on each candidate cloud to select the best one as the boundary cloud $\bm{P}_b$. To be specific, 
we 
select the best boundary-cloud $\bm{P}_{can}^b$ from the adversarial candidate clouds with the slightest distortion measured by distance metric $D(\bm{P}_s,\bm{P}_{can})$:
\begin{equation}\label{eq3}
\begin{aligned}
&\bm{P}_{can}^b = argmin_{\bm{P}_{can}^i} D(\bm{P}_s,\bm{P}_{can}^i)(i=1,2,...,T), \quad\\
&s.t.\ \varphi(\bm{P}_{can}^i)=1,
max_j(\left\| \bm{p}_{can,j}^i-\bm{p}_{s,j}\right\|_2)\le\varepsilon,
\end{aligned}
\end{equation}
where $\bm{P}_{can}^i$ denotes the $i$-th candidate-cloud, $\bm{p}_{can,j}^i$ denotes the $j$-th points in $\bm{P}_{can}^i$, $\bm{p}_{s,j}\in\bm{P}_s$ denotes the point with lowest distance from $\bm{p}_{can,j}^i$.
The distance measurement function $D(\cdot)$ is formulated as:
\begin{equation}\label{eq4}
\begin{aligned}
D(\bm{P}_s,&\bm{P}_{can})= D_{Chamfer}(\bm{P}_s,\bm{P}_{can}) \\ & +\hfill 
\gamma_1D_{Hausdorff}(\bm{P}_s,\bm{P}_{can})  +\gamma_2D_{Variance}(\bm{P}_s,\bm{P}_{can}),
\end{aligned}
\end{equation}
where $D_{Chamfer}$ and $D_{Hausdorff}$ measure the distance between two point clouds following \cite{wen2020geometry}. $D_{Variance}$ is the variance of distance between two point clouds. $\gamma_1$ and $\gamma_2$ are penalty parameters. 
We then project the $\bm{P}_{can}^b$ onto the decision boundary to obtain the boundary-cloud $\bm{P}_{b}$. Specifically, we conduct a binary search strategy \cite{nowak2008generalized} to move $\bm{P}_{can}^b$ towards $\bm{P}_{s}$ until reaching the decision boundary as:
\begin{equation}\label{eq5}
\begin{split}
\bm{P}_b&=\beta\bm{P}_s+(1-\beta)\bm{P}_{can}^b,\\
\end{split}
\end{equation}
where $\beta$ is the moving ratio in binary search. 


\subsection{Boundary-Cloud Optimization Along Decision Boundary}
The goal of boundary-cloud optimization is to further minimize the distance between the boundary cloud $\bm{P}_b$ and the source cloud $\bm{P}_s$ by moving $\bm{P}_b$ along the decision boundary to the optimal place so that the perturbation is smallest yet imperceptible while $\bm{P}_b$ keeps adversarial. 
Specifically, we utilize the iterative walking algorithm to optimize the point cloud $\bm{P}_b$ by two steps: \emph{1. Estimate updation gradient and move the point cloud along it}, \emph{2. Project the moved point cloud back to the decision boundary}.

\emph{1.Estimate updation gradient and move the point cloud along it.} As for initialization, we take $\bm{P}_b$ obtained in Section.\ref{section:3.3} as the initial boundary-cloud $\bm{P}_b^{(0)}$. For the next step, we denote $\bm{P}_b^{(t)}$ as the boundary-cloud obtained in the $t$-th walking iteration which is exactly on the decision boundary. We aim to estimate a gradient direction of $\bm{P}_b^{(t)}$ for improving the gap between the adversarial and the true class labels while preserving the geometric distance, so that we can make $\bm{P}_b^{(t)}$ more aggressive with small distortion and can further move it towards source cloud $\bm{P}_s$. Specifically, we utilize the \textit{point-wise walking} to conduct Monte Carlo method \cite{james1980monte} to obtain the estimated gradient vector $\triangledown S^{(t)}$ under the guidance of indicator function $\varphi(\cdot)$:
 \begin{equation}\label{eq6}
\triangledown S^{(t)}=\frac{1}{B} \sum_{i=1}^B \varphi(\bm{P}_b^{(t)}+v_i)v_i,
\end{equation}
where $v_i$ is the sampled move vectors obeying a normal distribution and $B$ is the number of vectors sampled in Monte Carlo method.
Then we move $\bm{P}_b^{(t)}$ along $\triangledown S^{(t)}$ with a step size $\xi$ by:
\begin{equation}\label{eq7}
\bm{P}_b^{temp}=\bm{P}_b^{(t)}+\xi\frac{\triangledown S^{(t)}}{\left\| \triangledown S^{(t)} \right\|_2}.
\end{equation}

\emph{2. Project the moved point cloud back to decision boundary.} 
After moving the point cloud along the estimated gradient, it will leave the decision boundary. Thus we conduct a binary search strategy to move $\bm{P}_b^{temp}$ towards $\bm{P}_s$ until reaching the decision boundary again. With multiple such walking iterations, we take the last $\bm{P}_b^{(t)}$ with the best optimization as the final adversarial point cloud.

So far, the optimization algorithm is complete, and we generate a desired adversarial point cloud $\bm{P}_b^{(t)}$.
However, as mentioned in Section.\ref{section3.2}, there remains the local optimum problem. To address it, in addition to the \textit{point-wise walking strategy}, we also conduct a \textit{spectrum-wise walking} to bring a great movement for cloud features and 
escape the convex area.
Specifically, we maintain other operations and solely replace the coordinate gradient walking in Eq.\ref{eq5} and Eq.\ref{eq6} with a spectrum gradient walking as:
 \begin{equation}\label{eq8}
\triangledown S^{(t)}=\frac{1}{B} \sum_{i=1}^B \varphi (IGFT(GFT(\bm{P}_b^{(t)})+u_i))u_i,
\end{equation}
\begin{equation}\label{eq9}
\bm{P}_b^{temp}=\bm{P}_b^{(t)}+\xi_{spe}\frac{\triangledown S^{(t)}}{\left\| \triangledown S^{(t)} \right\|_2},
\end{equation}
where $\xi_{spe}$ is a step size for spectrum walking.
By combining two walking operations for multi-step walking, the spectrum walking is able to perform large movement to jump out of local optima for a better optimization region, while the coordinate walking is able to perform slight movement to gradually fine-tune for getting the best optimization point in such region. The iterative walking algorithm is detailed in Algorithm.\ref{alg1}.
The optimized point cloud is our final adversarial sample.

\begin{algorithm}[h]
    \SetAlgoLined
    \caption{Joint Spectrum\&Coordinate Walking} \label{alg1}
    \LinesNumbered
    \KwIn{Boundary cloud $\bm{P}_b^{(0)}$, iteration step $R$, best optimized cloud list $Best=[]$}
    \KwOut{Adversarial point cloud $\bm{P}_{adv}$}
    $Best[0]=\bm{P}_b^{(0)}$\;
    \For{$i=1:R$}{ 
		conduct coordinate walking: 
            $\bm{P}_b^{(i)}\rightarrow\bm{P}_b^{(i+1)}$\;
		\eIf{$\bm{P}_b^{(i)}=\bm{P}_b^{(i+1)}$}{ 
                { 
                    conduct spectrum walking: $\bm{P}_b^{(i+1)}\rightarrow\bm{P}_b^{(i+1)}$\;
                }
		}{
                $Best\text{.append}(\bm{P}_b^{(i+1)})$\;
            }
    }
    Select $\bm{P}_{adv} \in Best$ with the smallest distance $D$\;
\end{algorithm}

\vspace{-16pt}
\section{Experiments}
\subsection{Dataset and 3D Models}
\noindent  \textbf{Dataset.} We use ModelNet40 \cite{wu20153d} in our experiments to evaluate the attack performance. ModelNet40 consists of 12,311 CAD models from
40 object categories, in which 9,843 models are intended
for training and the other 2,468 for testing.  Following the settings of previous work \cite{wen2020geometry,liu2022imperceptible,hu2022exploring}, we uniformly sample 1024 points from the surface of each object and scale them into a unit ball. For the adversarial point cloud attacks, we follow \cite{xiang2019generating,hu2022exploring} and randomly select 25 instances for each of 10 object categories in the ModelNet40 testing set, which can be well classified by the classifiers of interest.

\noindent  \textbf{3D Models.} We use three 3D networks as the victim models: PointNet \cite{qi2017pointnet}, PointNet++ \cite{qi2017pointnet++}, DGCNN \cite{wang2019dynamic}.
Experiments on other models can be found in supplementary.
\subsection{Implementation Details}
\noindent  \textbf{Attack Setup.}
We set K = 10 to build a K-NN graph to conduct Graph Fourier Transform (GFT). The weights of Hausdorff distance loss and variance loss e.g.$\gamma_1$ and $\gamma_2$ in Eq.\ref{eq4} are set to 2.0 and 0.5, respectively. For the spectrum fusion rate $\alpha$ in Eq.\ref{eq2}, the low frequencies weight $\alpha_{low}$ are set to 0.85 and the high frequencies weight $\alpha_{high}$ are set to 0.2. During an attack, we gradually reduce the fusion weight to distort the original point clouds more significantly until produced point clouds can confuse the victim model (point clouds from different object categories possess different sensitivity to fusion weights). 
We use $B = 50$ queries in the Monte Carlo algorithm to estimate the gradient. We conduct $R=200$ iteration rounds during boundary-cloud Optimization stage.
We set $\xi=\Vert \bm{P}_b^{(t)}-\bm{P}_s \Vert_2 / \sqrt{t}$ in Eq.\ref{eq7} and $\xi_{spe}=5.0$ in Eq.\ref{eq8}.

\noindent  \textbf{Evaluation Metrics.}
To quantitatively evaluate the effectiveness of our 3DHacker, we measure the perturbation size via three metrics: (1) L2-norm distance $D_{norm}$ \cite{cortes2012l2};  
(2) Chamfer distance $D_c$ \cite{fan2017point};
(3) Hausdorff distance $D_h$\cite{huttenlocher1993comparing}.
\subsection{Evaluation on Our 3DHacker Attack}

\begin{table*}[htbp]
\caption{Comparative results on the perturbation sizes of adversarial point clouds generated by different attack methods under 100\% ASR.}
\centering
\setlength{\tabcolsep}{1.0mm}{
\begin{tabular}{c|c|cc|ccc|ccc|ccc}
\hline
\multirow{2}*{Setting} & \multirow{2}*{Attack} & \multicolumn{2}{c}{Model Details} & \multicolumn{3}{|c}{PointNet \cite{qi2017pointnet}} & \multicolumn{3}{|c}{PointNet++ \cite{qi2017pointnet++}} & \multicolumn{3}{|c}{DGCNN \cite{wang2019dynamic}}\\
\cline{3-13}
~ & ~ & Para. & Logits &$D_h$ & $D_c$ & $D_{norm}$ &$D_h$& $D_c$ & $D_{norm}$ & $D_h$ & $D_c$ & $D_{norm}$ \\
\hline
\multirow{7}*{White-Box}
&FGSM \cite{yang2019adversarial}&$\checkmark$&$\checkmark$&0.1853&0.1326&0.7936 &0.2275&0.1682&0.8357 &0.2506&0.189&0.8549 \\
&PGD \cite{madry2017towards}&$\checkmark$&$\checkmark$&0.1322&0.1329&0.7384&0.1623&0.1138&0.7596&0.1546&0.1421&0.7756 \\
&AdvPC \cite{hamdi2020advpc}&$\checkmark$&$\checkmark$&0.0343&0.0697&0.6509&0.0429&0.0685&0.6750&0.0148&0.0623&0.6304\\
&3D-ADV$^p$ \cite{xiang2019generating}&$\checkmark$&$\checkmark$&0.0105&0.0003&0.5506 &0.0381&0.0005&0.5699&0.0475&0.0005&0.5767\\
&LG-GAN \cite{zhou2020lg}&$\checkmark$&$\checkmark$&0.0362&0.0347&0.7184&0.0407&0.0238&0.6896&0.0348&0.0119&0.8527\\
&GeoA$^3$ \cite{wen2020geometry}&$\checkmark$&$\checkmark$&0.0175&0.0064&0.6621&0.0357&0.0198&0.6909&0.0402&0.0176&0.7024\\
&SI-Adv$^w$ \cite{huang2022shape}&$\checkmark$&$\checkmark$&0.0204&0.0002&0.7614&0.0310&0.0004&1.2830&0.0127&0.0006&1.1120\\

\hline
Black-Box&SI-Adv$^b$ \cite{huang2022shape}&$\times$&$\checkmark$&0.0431&0.0003&0.9351&0.0444&0.0003&1.0857&0.0336&0.0004&0.9081\\
\hline
\multirow{2}*{\tabincell{c}{Hard-Label\\ Black-Box}}
&\multirow{2}*{Ours}&\multirow{2}*{$\times$}&\multirow{2}*{$\times$}&\multirow{2}*{\textbf{0.0136}}&\multirow{2}*{\textbf{0.0017}}&\multirow{2}*{\textbf{0.8561}}&\multirow{2}*{\textbf{0.0245}}&\multirow{2}*{\textbf{0.0023}}&\multirow{2}*{\textbf{0.9324}}&\multirow{2}*{\textbf{0.0129}}&\multirow{2}*{\textbf{0.0026}}&\multirow{2}*{\textbf{0.9030}}\\
&&&&&&&&&&\\ 
\hline
\end{tabular}}
\vspace{-6pt}
\label{tab:1}
\end{table*}

\begin{table*}[htbp]
\caption{Resistance of the black-box attacks on defended point cloud models.}
\label{table:2}
\centering
\setlength{\tabcolsep}{0.6mm}{
\begin{tabular}{c|c|cccc|cccc|cccc}
\hline
\multirow{2}*{Defense} & \multirow{2}*{Attack} &\multicolumn{4}{|c}{PointNet \cite{qi2017pointnet}} & \multicolumn{4}{|c}{PointNet++ \cite{qi2017pointnet++}} & \multicolumn{4}{|c}{DGCNN \cite{wang2019dynamic}}\\
\cline{3-14}
~ & ~ & ASR(\%) & $D_h$ & $D_c$& $D_{norm}$  & ASR(\%) & $D_h$ & $D_c$ & $D_{norm}$ &ASR(\%) & $D_h$ &$D_c$& $D_{norm}$ \\
\hline
\multirow{2}*{SOR\cite{zhou2019dup}} & SI-Adv$^b$ \cite{huang2022shape} &89.7 & 0.0420  & \textbf{0.0009}& 3.0193
& 78.9 &0.0436 & \textbf{0.0025}& 1.3843 
& 72.0 & 0.0341 &\textbf{0.0009}& 1.6480\\ 
~ & Ours&\textbf{90.4} & \textbf{0.0100} &0.0023& \textbf{1.2486} 
& \textbf{82.7} & \textbf{0.0218} &0.0043& \textbf{1.3759} 
& \textbf{85.4} & \textbf{0.0124} &0.0031& \textbf{1.2387}\\ 
\hline
\multirow{2}*{Drop(30\%)} & SI-Adv$^b$ \cite{huang2022shape} &96.9 & 0.0426 &\textbf{0.0003}& 1.3680 
& 70.1 & 0.0473 &\textbf{0.0023}& 1.4538 
& 71.2 & 0.0400 &\textbf{0.0004}& \textbf{0.8598}\\  
~ & Ours& \textbf{97.2} & \textbf{0.0179} &0.0016& \textbf{0.8391}
& \textbf{71.3} & \textbf{0.0298} &0.0031& \textbf{1.2810} 
& \textbf{78.5} & \textbf{0.0195} &0.0033& 1.1742\\ 
\hline
\multirow{2}*{Drop(50\%)} & SI-Adv$^b$ \cite{huang2022shape} &93.6 & 0.0420 &\textbf{0.0002}&  1.3844 
& 67.6 &0.0501 &\textbf{0.0013}& 1.9193 
& 75.2 & 0.0358 &\textbf{0.0004}& \textbf{0.6992}\\  
~ & Ours&\textbf{95.4} & \textbf{0.0182} &0.0023& \textbf{0.8328}
& \textbf{77.4} & \textbf{0.0285} &0.0032& \textbf{1.4735}
& \textbf{76.8} & \textbf{0.0172} &0.0036& 1.2914\\ 
\hline
\end{tabular}}
\vspace{-10pt}
\end{table*}

\noindent  \textbf{Comparison with existing methods.} To investigate the effectiveness of our attack, we perform several existing white-box adversarial attacks and one black-box adversarial attack for quantitative comparison as shown in Table~\ref{tab:1}.
Table~\ref{tab:1} shows that our 3DHacker achieves smaller perturbation sizes than the black-box model and achieves very competitive results with white-box models.
Since our 3DHacker conducts global perturbations to origin point clouds which possess a strong potential to confuse the victim models with a structure distortion, this global perturbations produce a higher Chamfer distance $D_c$ compared with 3D-ADV$^p$ and SI-Adv because $D_c$ measures the average squared distance between each adversarial point and its nearest original point and we modify all the points leading to a large sum of displacements. Instead, attacking by modifying a few points in 3D-ADV$^p$ and SI-Adv has advantage in $D_c$ because most of the distance is equal to 0. However, 3DHacker performs better in $D_h$ since we conduct relatively average perturbations to point cloud which does not count on a few outliers to confuse the victim models, leading to imperceptible and having the potential to bypass the outlier detection defense.

\begin{figure}[ht]
	\centering
	\vspace{-10pt}
 \includegraphics[width=0.47\textwidth]{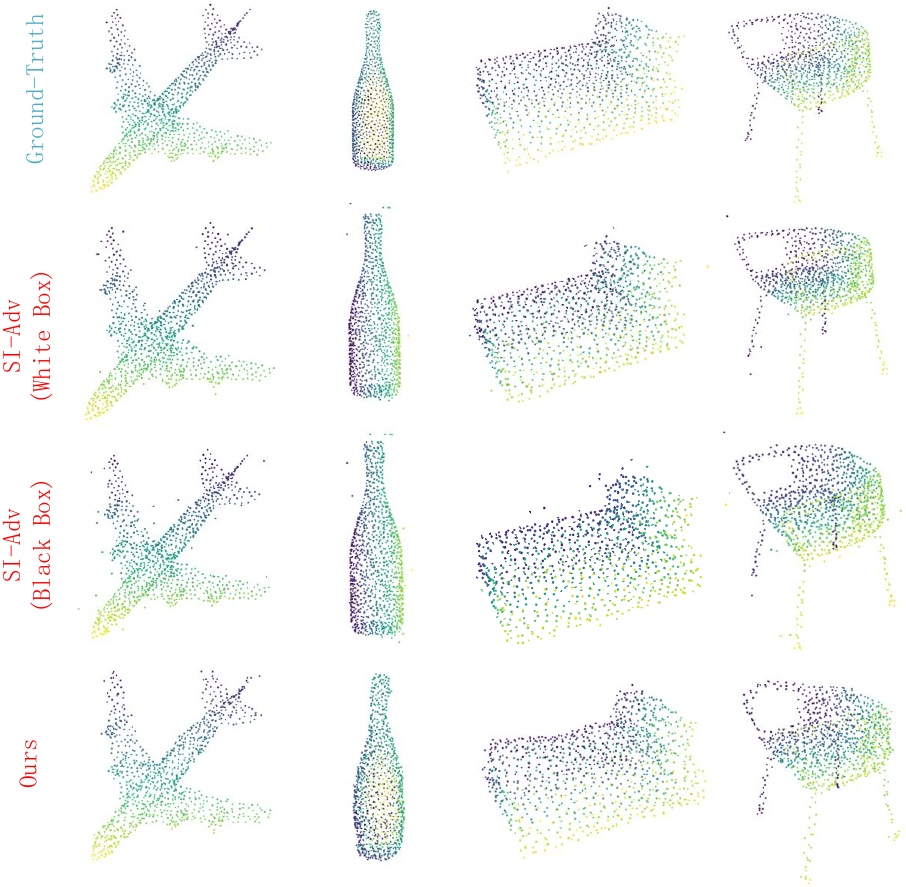}
	\caption{Visualization results of adversarial samples.}
 \vspace{-10pt}
	\label{fig_3}
\end{figure}

\noindent  \textbf{Visualization results.} We provide visualization on adversarial samples generated by our 3DHacker, SI-Adv$^w$\cite{huang2022shape} (white box attack) and SI-Adv$^b$\cite{huang2022shape} (black box attack) in Figure~\ref{fig_3}. Our 3DHacker can alleviate the outlier point problems and produce more imperceptible adversarial samples.

\noindent  \textbf{Resistance to Defenses.}
To evaluate the robustness of our 3DHacker against different adversarial defenses, we conduct the experiments on two widely used defense methods: Statistical Outlier Removal(SOR) \cite{zhou2019dup} and Simple Random Sampling(SRS) \cite{yang2019adversarial}. Following the defense experiments setting on SRS in \cite{huang2022shape}, we conduct the Simple Random Sampling by randomly dropping 30\% and 50\% of input points respectively. As shown in Table~\ref{table:2}, (1) Our 3DHacker can achieve a higher attack success rate than SI-Adv$^b$ when attacking the model protected by SOR. This is because our method alleviates the outlier point problems and selects the best adversarial samples with the smallest perturbations, while SI-Adv$^b$ still suffers from the perturbed point of outlier in the sharp component.
(2) As for SRS defense, our 3DHacker still achieves a better attack than SI-Adv$^b$ as we generate the adversarial sample with high similarity to the original one in both geometric topology and local point distributions.
Overall, our 3DHacker is much more robust to existing defense strategies.

\subsection{Ablation Study}
\begin{figure}[t!]
	\centering
	\includegraphics[width=0.47\textwidth]{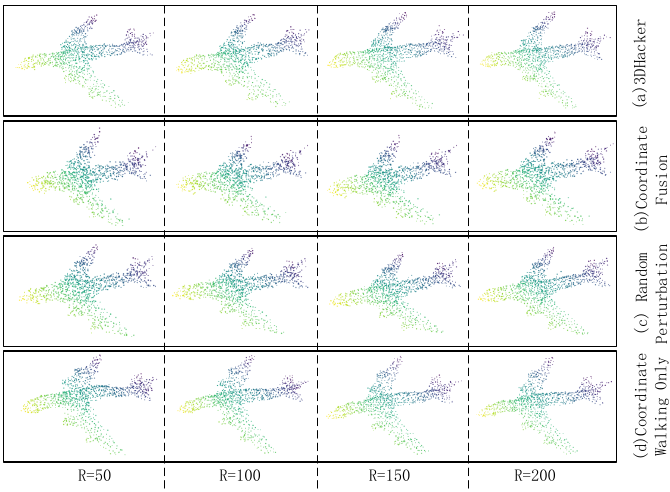}
 \vspace{-8pt}
	\caption{Visualization on different ablations.}
 \vspace{-6pt}
	\label{fig_4}
\end{figure}

\noindent  \textbf{Investigation on different strategies for boundary-cloud generation.} 
To verify the effects of our spectrum fusion method in boundary-cloud generation stage, we conduct the experiments by replacing the spectrum fusion method with different strategies while maintaining the latter procedure and settings in boundary-cloud optimization stage the same. Specifically, two general strategies are compared: traditional coordinate fusion (which fuses source-cloud and target-cloud in coordinate space with proper fusion rate) and simple random perturbation (which directly add point-wise noise to source-cloud to reach the decision boundary). As shown in Table \ref{table:3}, our spectrum fusion achieves the smallest perturbations than other strategies in all metrics, this is because: (1) coordinate fusion will destroy to geometric structure by averaging different shapes of 3D objects; (2) random perturbation will lead to outliers and uneven point distribution without geometric awareness. Their visualized adversarial samples are shown in Figure~\ref{fig_4} (a,b,c), where our samples are more imperceptible.

\begin{table}[t!]
\caption{Investigation on different strategies in the boundary-cloud generation stage. Victim model: PointNet.}
\label{table:3}
\centering
\setlength{\tabcolsep}{1.6mm}{
\begin{tabular}{cccc}
\hline
Generation Strategy  & $D_h$ & $D_c$ & $D_{norm}$\\
\hline
Spectrum Fusion& \textbf{0.0136} & \textbf{0.0017}& \textbf{0.8561} \\
Coordinate Fusion &0.0275 & 0.0038& 1.4379\\
Random Perturbation& 0.0256 & 0.0023& 0.9673 \\
\hline
\end{tabular}}
\vspace{-6pt}
\end{table}

\begin{table}[t!]
\caption{The effect of different walking methods in the boundary-cloud optimization stage. Victim model: PointNet.}
\label{table:4}
\centering
\setlength{\tabcolsep}{1.6mm}{
\begin{tabular}{ccccc}
\hline
\multirow{2}*{\tabincell{c}{Coordinate\\ Walking}} & \multirow{2}*{\tabincell{c}{Spectrum\\ Walking}} &\multirow{2}*{$D_h$}&\multirow{2}*{$D_c$}&\multirow{2}*{$D_{norm}$}\\
&\\
\hline
$\checkmark$& $\checkmark$ & \textbf{0.0136} & \textbf{0.0017}& \textbf{0.8561} \\
$\checkmark$& $\times$ &0.0198 & 0.0047 & 1.5827\\
$\times$ & $\checkmark$ &0.0598 & 0.0109 & 3.1746\\
\hline
\end{tabular}}
\vspace{-6pt}
\end{table}

\begin{table}[t!]
\caption{Analysis on Iteration Round $R$. Victim model: PointNet.}
\label{table:5}
\centering
\setlength{\tabcolsep}{1.6mm}{
\begin{tabular}{cccc}
\hline
Iteration Rounds R  & $D_h$ & $D_c$ & $D_{norm}$\\
\hline
$R=100$& 0.0185 & 0.0024& 1.1613 \\
$R=150$& 0.0152 & 0.0022& 0.9470 \\
$R=200$& 0.0136 & 0.0017& 0.8561 \\
$R=250$& \textbf{0.0131} & \textbf{0.0016}& \textbf{0.8487} \\
\hline
\end{tabular}}
\vspace{-6pt}
\end{table}

\noindent  \textbf{The effect of different walking methods for boundary-cloud optimization.} In the boundary-cloud optimization stage, we design a spectrum walking method in addition to the coordinate one to jump out of the local optimum. 
To investigate the effect of each walking strategy, as shown in Table~\ref{table:4}, we remove one of them to conduct the ablations for comparison. From this table, without spectrum walking, the attack process is easily trapped into the local optimum, leading to larger perturbations. Without coordinate walking, as shown in Figure~\ref{fig_4} (d), it is hard to measure the point-wise imperceptibility for optimization, thus achieving the worst performance. By utilizing both of them, our model can preserve both high imperceptibility and geometric smoothness.

\begin{table}[t!]
\caption{Sensitivity analysis on Numbers of selected samples $B$ in Monte Carlo algorithm. Victim model: PointNet.}
\label{table:6}
\centering
\setlength{\tabcolsep}{1.6mm}{
\begin{tabular}{cccc}
\hline
selected samples $B$ & $D_h$ & $D_c$& $D_{norm}$ \\
\hline
$B=10$& 0.0164 & 0.0023& 1.3725 \\
$B=30$& 0.0141 & 0.0018& 0.9636 \\
$B=50$& \textbf{0.0136} & \textbf{0.0017}& \textbf{0.8561} \\
$B=70$& 0.0139 & \textbf{0.0017}& 0.9073 \\
\hline
\end{tabular}}
\vspace{-10pt}
\end{table}

\noindent  \textbf{Sensitivity on the iteration rounds $R$.} As shown in Table.\ref{table:5}, we conduct the ablation on the iteration rounds $R$ of the boundary-cloud optimization. Our model achieves the best performance when $R$ is set to 250. However, the model with $R=250$ is slightly better than the model with $R=200$, but leads to much more time consumption. To balance both the performance and time cost, we choose $R=200$ in our all experiments. Visualization on adversarial point clouds of different $R$ is shown in Figure~\ref{fig_4}. 

\noindent  \textbf{Sensitivity on number $B$ in Monte Carlo algorithm.}
As shown in Table~\ref{table:6}, we conduct the ablation on the number $B$ of selected samples in Monte Carlo algorithm. It shows that we achieve the smallest perturbation when $B=50$.

\begin{figure}[t!]
	\centering
	\includegraphics[width=0.47\textwidth]{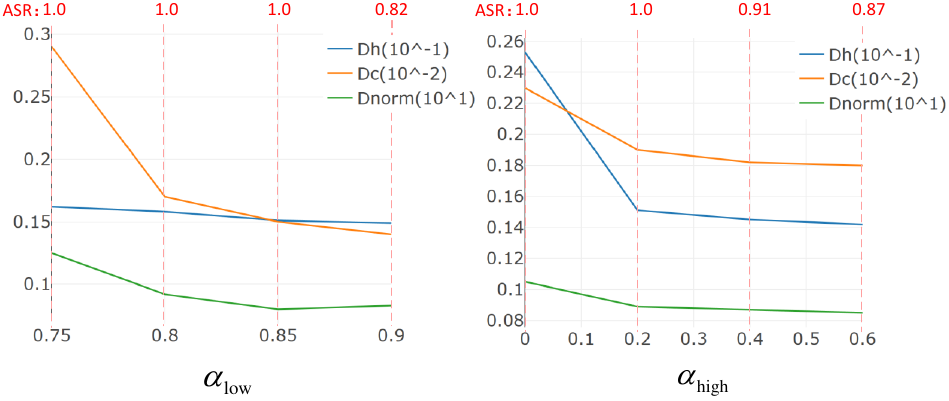}
 \vspace{-6pt}
	\caption{Influence of different spectrum fusion rates.}
 \vspace{-10pt}
	\label{fig_5}
\end{figure}

\noindent  \textbf{Influence of different spectrum fusion rate.}
The spectrum fusion rate $\alpha_{low}$ and $\alpha_{high}$ decide the fusion weights of rough shape and fine details in Eq.\ref{eq2}, respectively. 
Here, we modify $\alpha_{low}$ and $\alpha_{high}$ to analyze their influence on attack quality. In Figure~\ref{fig_5}, by balancing perturbation size and attack success rate, we set $\alpha_{low}=0.85,\alpha_{high}=0.2$.

\section{Conclusion}
We introduce a new and challenging 3D attack setting, \textit{i.e.}, attacking point clouds with black-box hard labels. To address this practical setting, we propose a novel attack method called 3DHacker based on our decision boundary algorithm, which adopts spectrum fusion to generate boundary clouds with high imperceptibility and employs an additional spectrum walking strategy to move the boundary clouds along the decision boundary for further optimization with trivial perturbations. Experiments validate both effectiveness and robustness of our 3DHacker.

\noindent \textbf{Acknowledgements.}
This work is funded by the National Natural Science Foundation of China (No.61972009, No.61972448, No.61972449), and the National Key Research and Development Program (No.2022YFB4501300).

{\small
\bibliographystyle{ieee_fullname}
\bibliography{egbib}
}


\begin{table*}[t!]
\caption{Comparative results on the perturbation sizes of different methods for adversarial point clouds. \textbf{\textit{Our setting is harder to attack.}}}
\centering
\setlength{\tabcolsep}{1.0mm}{
\begin{tabular}{c|c|cc|ccc|ccc|ccc}
\hline
\multirow{2}*{Setting} & \multirow{2}*{Attack} & \multicolumn{2}{c}{Model Details} & \multicolumn{3}{|c}{PAConv \cite{xu2021paconv}} & \multicolumn{3}{|c}{SimpleView \cite{goyal2021revisiting}} & \multicolumn{3}{|c}{CurveNet \cite{xiang2021walk}}\\
\cline{3-13}
~ & ~ & Para. & Logits &$D_h$ & $D_c$ & $D_{norm}$ &$D_h$& $D_c$ & $D_{norm}$ & $D_h$ & $D_c$ & $D_{norm}$ \\
\hline
White-Box&SI-Adv$^w$ \cite{huang2022shape}&$\checkmark$&$\checkmark$&0.0097&0.0004&0.6920&0.0256&0.0014&2.1522&0.0199&0.0006&0.9803\\
\hline
Black-Box&SI-Adv$^b$ \cite{huang2022shape}&$\times$&$\checkmark$&0.0449&0.0004&1.3386&0.0469&0.0010&1.8754&0.0453&0.0004&1.4336\\
\hline
Hard-Label
&\multirow{2}*{Ours}&\multirow{2}*{$\times$}&\multirow{2}*{$\times$}&\multirow{2}*{0.0046}&\multirow{2}*{0.0014}&\multirow{2}*{0.9444}&\multirow{2}*{0.0136}&\multirow{2}*{0.0029}&\multirow{2}*{1.6150}&\multirow{2}*{0.0125}&\multirow{2}*{0.0022}&\multirow{2}*{1.2332}\\
Black-Box&&&&&&&&&&\\ 
\hline
\end{tabular}}
\label{tab:1}
\end{table*}

\begin{table*}[t!]
\caption{Resistance of the black-box attacks on defended point cloud models.}
\label{table:2}
\centering
\setlength{\tabcolsep}{1.6mm}{
\begin{tabular}{c|c|ccc|ccc|ccc}
\hline
\multirow{2}*{Defense} & \multirow{2}*{Attack} &\multicolumn{3}{|c}{PAConv \cite{xu2021paconv}} & \multicolumn{3}{|c}{SimpleView \cite{goyal2021revisiting}} & \multicolumn{3}{|c}{CurveNet \cite{xiang2021walk}}\\
\cline{3-11}
~ & ~ & ASR(\%) & $D_h$ & $D_{norm}$ & ASR(\%) & $D_h$ & $D_{norm}$ &ASR(\%) & $D_h$ & $D_{norm}$ \\
\hline
\multirow{2}*{SOR \cite{zhou2019dup}} & SI-Adv$^b$ \cite{huang2022shape} 
&94.4 & 0.0359  & 1.9640
& 95.2 &0.0375 & 3.1333 
& 88.8 & 0.0351 & 2.5402\\ 
~ & Ours&95.5 & 0.0028 & 0.6744 
& 93.6 & 0.0083 & 1.0873 
& 89.2 & 0.0095 & 1.1752\\ 
\hline
\multirow{2}*{Drop(30\%)} & SI-Adv$^b$ \cite{huang2022shape} 
&73.6 & 0.0402 &  1.1979
& 56.8 & 0.0411 &  1.2577
& 71.2 & 0.0400 & 1.4630\\  
~ & Ours& 95.2 & 0.0061 & 0.8290
& 91.2 & 0.0092 &  0.9638
& 82.5 & 0.0157 & 0.8598\\ 
\hline
\multirow{2}*{Drop(50\%)} & SI-Adv$^b$ \cite{huang2022shape} 
&84.8 & 0.0390 &   0.8537
& 68.8 & 0.0368 & 0.9119
& 79.2 & 0.0392 & 1.1759\\  
~ & Ours& 93.8 & 0.0136 & 0.7261
& 97.6 & 0.0066 & 0.7570 
& 83.4 & 0.0186 & 0.7558\\ 
\hline
\end{tabular}}
\end{table*}

\appendix

\section*{Appendix}

In the supplementary material, we first implement our attack method on more victim models for attack performance comparison, then we provide corresponding defense comparison to validate the robustness of our attack. After that, we provide more visualization results on the adversarial examples generated by different 3D attackers on different victim models. Finally, we provide more details of our proposed spectrum iterative walking strategy. 

\section{Attack Performance on More Victim Models}

To investigate the effectiveness and generalization-ability of our attack, we perform our 3DHackker on more victim models, \textit{i.e.}, PAConv \cite{xu2021paconv}, SimpleView \cite{goyal2021revisiting}, and CurveNet \cite{xiang2021walk}. For comparison, we select the SOTA attack method SI-Adv \cite{huang2022shape} in both white- and black-box settings. 
\textit{Note that, our 3DHacker is the first 3D adversarial attack in more challenging hard-label black-box setting, which is much harder to achieve success since it has no information of model details (white-box) and output logits (black-box).}
As shown in Table~\ref{tab:1}, our 3DHacker achieves smaller perturbation sizes than the black-box SI-Adv$^b$ model and achieves very competitive results with the white-box SI-Adv$^w$ model. 
Overall, our 3DHacker achieves the lowest perturbation $D_h$ in all three victim models, demonstrating the effectiveness of our 3DHacker.

\section{Defense on More Victim Models}
To evaluate the robustness of our 3DHacker compared to SI-Adv$^b$ \cite{huang2022shape}, we also conduct the defense methods Statistical Outlier Removal (SOR) \cite{zhou2019dup} and Simple Random Sampling (SRS) \cite{yang2019adversarial}) on corresponding adversarial examples generated on PAConv \cite{xu2021paconv}, SimpleView \cite{goyal2021revisiting}, and CurveNet \cite{xiang2021walk}. As shown in Table~\ref{table:2}, 
(1) As for the defense method SOR, our 3DHacker can achieve a higher attack success rate than SI-Adv$^b$ on all three victim models. 
(2) As for the SRS defense, our 3DHacker still achieves a better attack performance than SI-Adv$^b$ as we generate the adversarial sample with high similarity to the original one in both geometric topology and local point distributions. 
(3) Our adversarial samples achieve the lowest perturbations with a much higher attack success rate when attacking the model protected by defenses.
Overall, compared to the previous best attack method SI-Adv, our 3DHacker is much more robust to existing defense strategies.

\section{More Qualitative Results}
\begin{figure*}[t!]
	\centering
	\includegraphics[width=\textwidth]{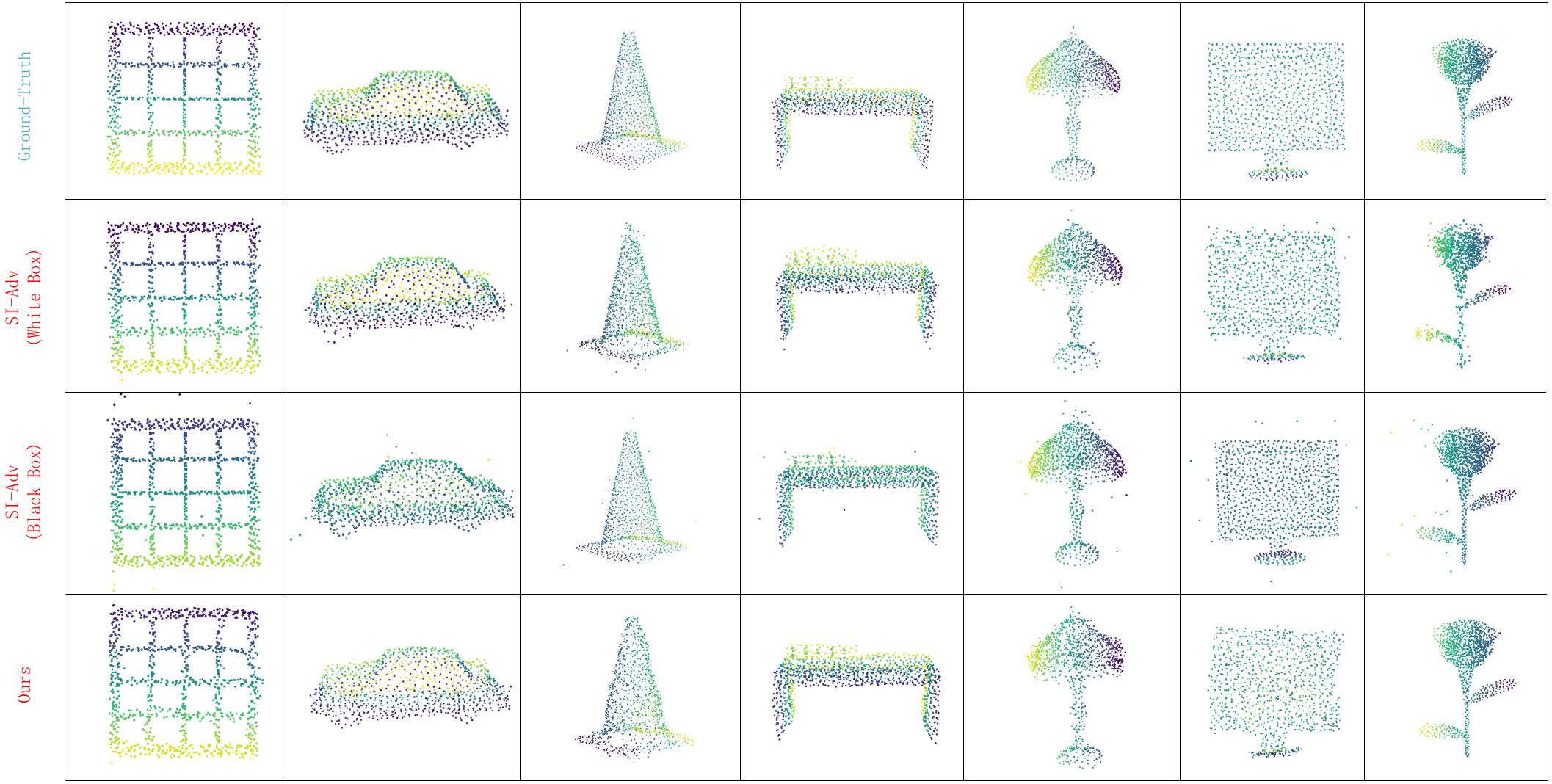}
	\caption{Visualization results of adversarial samples generated by different attack methods on PAConv model.}
 \vspace{-10pt}
	\label{sfig_1}
\end{figure*}

\begin{figure*}[t!]
	\centering
	\includegraphics[width=\textwidth]{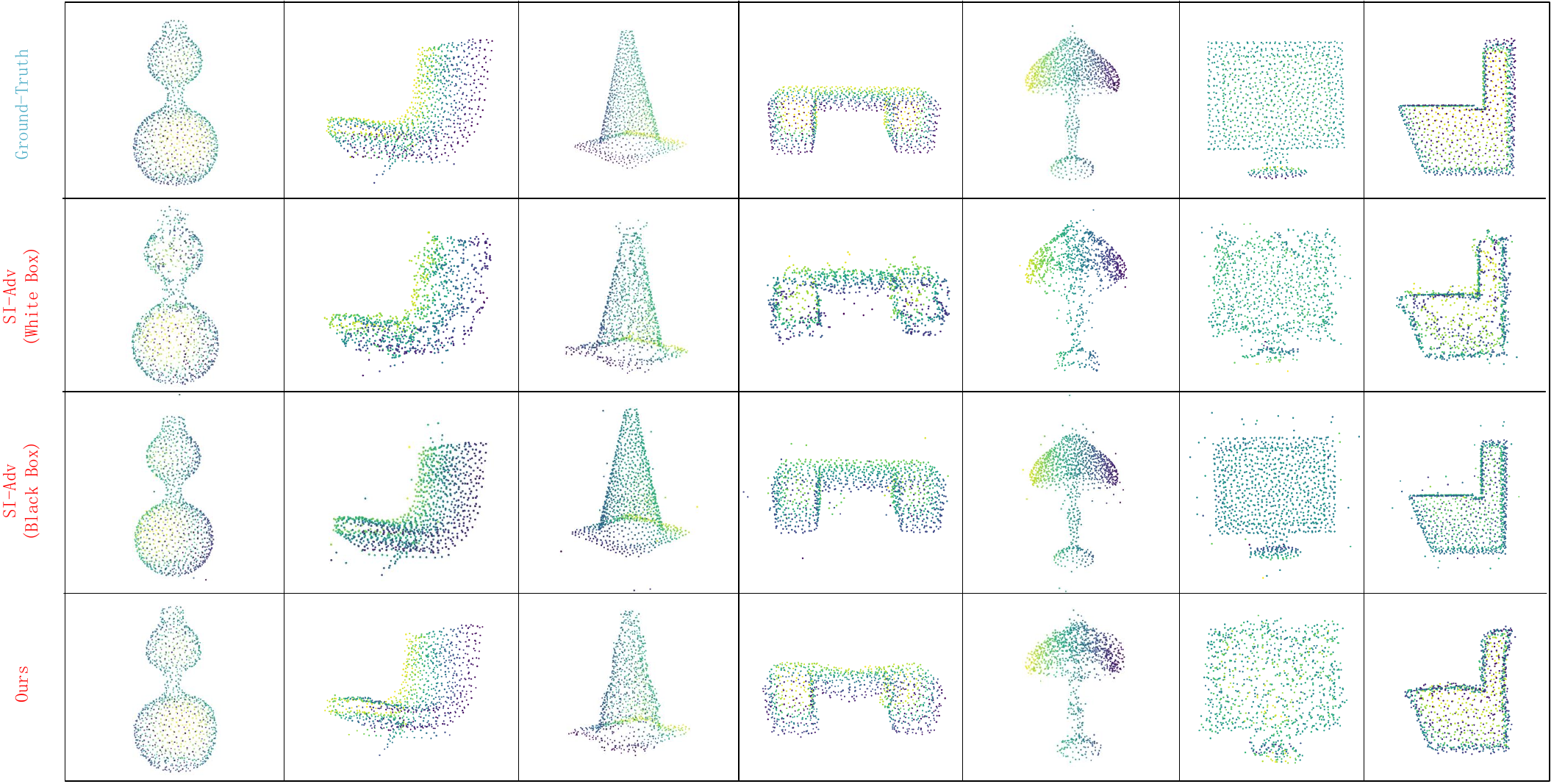}
	\caption{Visualization results of adversarial samples generated by different attack methods on SimpleView model.}
	\label{sfig_2}
\end{figure*}

\begin{figure*}[t!]
	\centering
	\includegraphics[width=\textwidth]{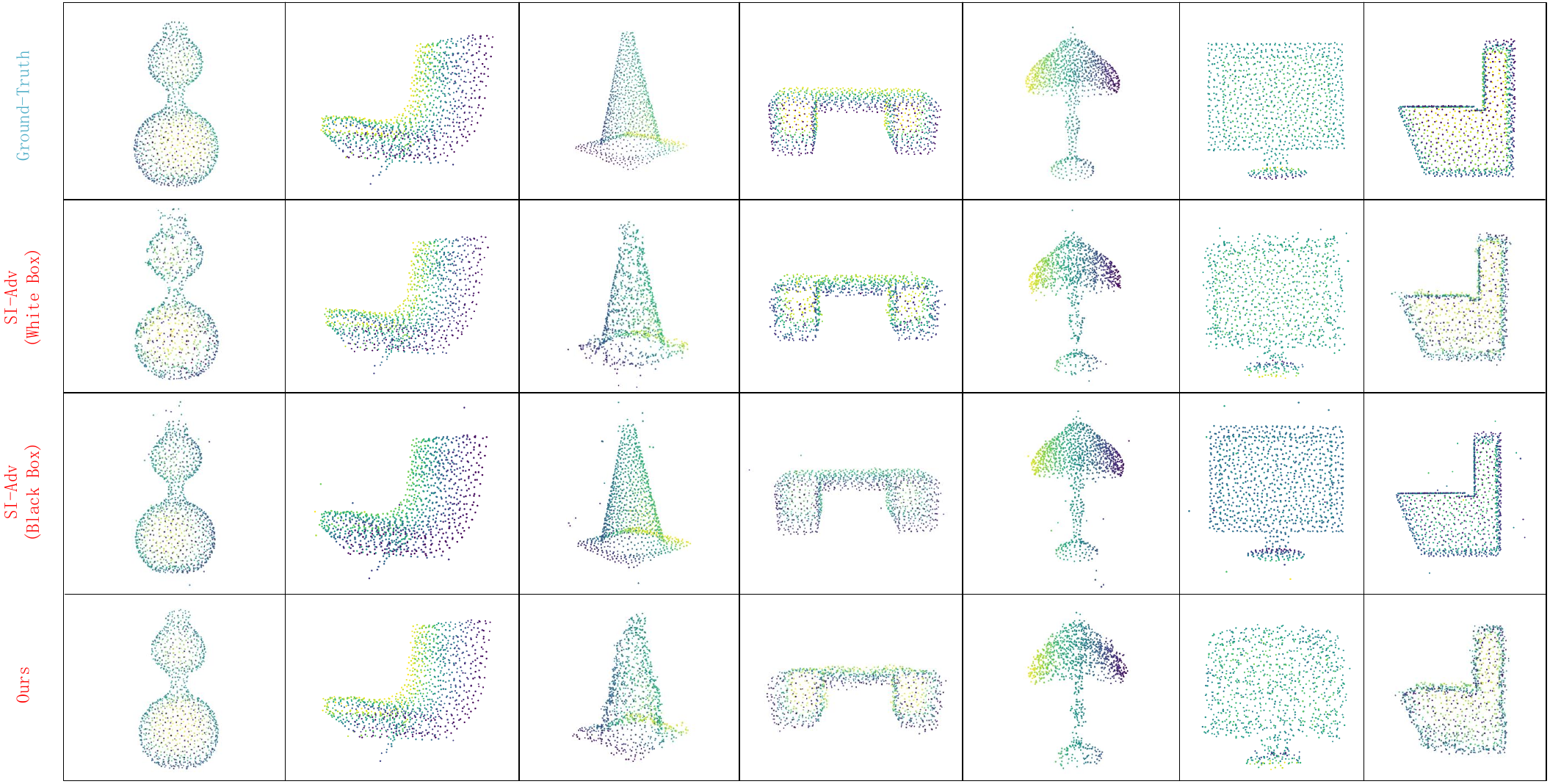}
	\caption{Visualization results of adversarial samples generated by different attack methods on DGCNN model.}
 \vspace{-10pt}
	\label{sfig_3}
\end{figure*}

To further demonstrate the effectiveness of our method on other point clouds of different object categories, we expand the visualization experiment that provides visualization on adversarial samples generated by our 3DHacker, SI-Adv$^w$\cite{wen2020geometry} (white box attack) and SI-Adv$^b$\cite{huang2022shape} (black box attack) as shown in Figure~\ref{sfig_1}, Figure~\ref{sfig_2} and Figure~\ref{sfig_3}. It shows that previous white- and black-box attackers easily lead to outlier problems and uneven distributions. Moreover, they require more knowledge of the model details (parameters or output logits) during the generation process of adversarial samples. Compared to them, our hard-label setting only accesses the output label of the model and is harder to achieve successful attack. Even though, as shown in the figures, our 3DHackker can alleviate the outlier point problems and produce more imperceptible adversarial samples.

\begin{figure*}[t!]
	\centering
	\includegraphics[width=\textwidth]{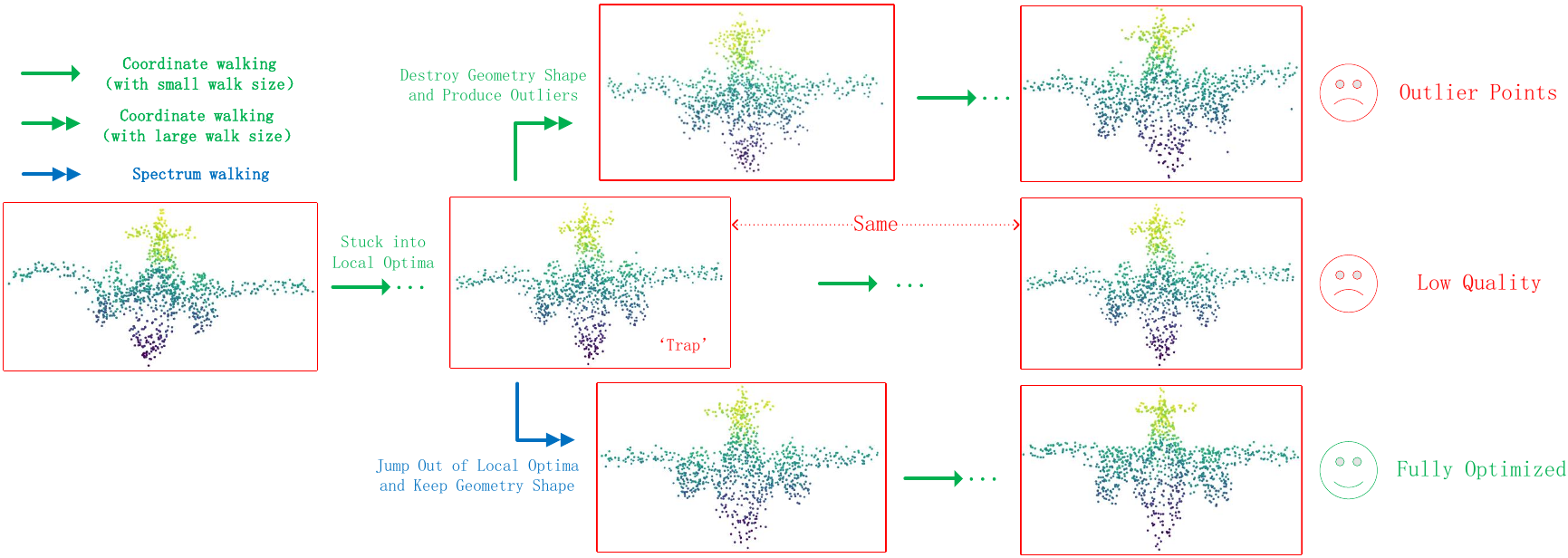}
	\caption{Visualization on the different combinations of coordinate and spectrum walking strategies for optimizing adversarial point cloud.}
 \vspace{-10pt}
	\label{sfig_4}
\end{figure*}

\section{More Details of Spectrum Walking}
As mentioned in Section 3.4 of the main paper, in addition to the general coordinate walking, we design a spectrum-wise walking strategy in the boundary-cloud optimization stage to jump out the local optimum for producing a better optimized adversarial point cloud. 
Here, we first provide more details of this local optimum problem, and then explain why our spectrum walking strategy work. 

\noindent \textbf{Details of local optimum.}
By only utilizing the coordinate walking to move the boundary cloud along the decision boundary, the optimization process may stop earlier and stuck into a local concave of the decision boundary. 
For example, if we design an optimization process with 200 iterations for each boundary cloud, some optimized point clouds may keep constant at the beginning. 
This is because the adversarial point cloud has a chance to fall into a `trap' due to the concave-convex of the decision boundary, where a further small walking step in arbitrary direction is likely to change the classification result of the victim model to the ground-truth label of benign cloud, thus it is hard to estimate a gradient direction of coordinate walking while keeping adversarial in the next iteration. 
We call this phenomenon as the local optimum problem, and the boundary cloud falling into the `trap' may possess low quality.
To this end, in addition to the data domain, we need to explore additional knowledge in other latent spaces to adjust the point cloud geometry without losing its antagonism.

\noindent \textbf{Why spectrum walking work?}
To overcome such local optimum, the adversarial point cloud needs to walk a long step when falling into a `trap'. A general intuition is to increase the coordinate walking size, however, directly utilizing a large coordinate walking step will produce outliers that are hard to be eliminated in the following iterations, since the outliers contribute more to the adversarial performance than ordinary points. 
Therefore, we design a spectrum walking strategy in the spectral domain instead of the simple data domain, which not only can preserve high-quality geometric shape of the point cloud, but also has the potential to keep its latent adversarial characteristics during the spectrum walking optimization.
Moreover, unlike the coordinate-wise strategy that adds point-wise offsets for walking, walking in the spectral domain is to search trivial offsets of the spectrum frequency and will not lead to the data-domain problems of changing classification results and destroying the shape.
Therefore, spectrum walking is effective enough to help to jump out of the local optimum and avoid the outlier problems.
However, only utilizing the spectrum walking is not decision-boundary awareness, validated in Table 4 of the main paper.
Overall, by jointly utilizing coordinate and spectrum walking strategies, we can take advantage of both of them, and optimize the best adversarial point cloud along the decision boundary.
Figure~\ref{sfig_4} also illustrates the effectiveness of the joint coordinate-spectrum walking strategy.

\section{Other experimental results}
\noindent \textbf{Running time.} We conduct running time experiments to evaluate the attack efficiency of our 3DHacker. As shown in Table~\ref{tab:time}, our running time is competitive to the black-box model since our optimization steps can be efficiently achieved. The white-box model is most time-consuming since it needs complicated backpropagation through the victim model.

\begin{table}[h!]
\small 
    \centering
    \scalebox{0.9}{
    \begin{tabular}{c|cccc}
    \hline
    Method & PointNet & DGCNN & CurveNet & PAConv \\ \hline
    SI-ADV$^w$ & 1.32s & 3.87s & 21.53s & 2.18s\\ 
    SI-ADV$^b$ & 0.58s & 1.25s & 8.77s & 0.31s \\ 
    Ours & 1.16s & 2.18s & 10.60s & 1.09s\\ \hline
    \end{tabular}}
    \caption{\textbf{Average time} for each adversarial point cloud generation.}
    \label{tab:time}
    \vspace{-10pt}
\end{table}

\noindent \textbf{Comparison on hard-label settings.}
Since existing 3D attacks rely on either model parameters or output logits, they can not be adapted to hard-label setting. Therefore, we re-implement two 2D hard-label settings into 3D domain for comparison. In Table~\ref{tab:hard}, our method performs much better.

\begin{table}[t!]
\small
    \centering

    \scalebox{0.9}{
    \setlength{\tabcolsep}{1.0mm}{
    \begin{tabular}{c|ccc|ccc}
    \hline
    \multirow{2}*{Method} & \multicolumn{3}{c|}{PointNet} & \multicolumn{3}{c}{DGCNN} \\
    & $D_h$ & $D_c$ & $D_{norm}$ & $D_h$ & $D_c$ & $D_{norm}$ \\ \hline
    Chen \textit{et al.} 2020 & 0.1284&0.0695&1.1784&0.1291&0.0493&0.9827 \\
    Li \textit{et al.} 2021 &0.0814&0.0445&1.0863&0.0892&0.0505&1.1338\\
    Ours & \textbf{0.0136} & \textbf{0.0017} & \textbf{0.8561} & \textbf{0.0129} & \textbf{0.0026} & \textbf{0.9030} \\ \hline
    \end{tabular}}}
    \caption{Comparison on the same \textbf{hard-label setting}.}
    \vspace{-5pt}
    \label{tab:hard}
\end{table}

\noindent \textbf{Experiments on \textbf{ShapeNetPart dataset} for other victim models.}
We conduct additional experiments on novel victim point cloud classification models \cite{zhao2021point,yu2022point} and achieve remarkable performance similar to the results performed in main body. Our 3DHacker produces a higher Chamfer distance $D_c$ because we modify all the points leading to a large sum of displacements. However, it performs better in $D_h$ since we conduct relatively average perturbations to point cloud which does not count on a few outliers to confuse the victim models, leading to imperceptible and having the potential to bypass the outlier detection defense.

\begin{table}[t!]
\small
    \centering

    \scalebox{0.9}{
    \setlength{\tabcolsep}{1.4mm}{
    \begin{tabular}{c|ccc|ccc}
    \hline
    \multirow{2}*{Method} & \multicolumn{3}{c|}{PointTransformer [B]} & \multicolumn{3}{c}{Point-BERT [C]} \\
    & $D_h$ & $D_c$ & $D_{norm}$ & $D_h$ & $D_c$ & $D_{norm}$ \\ \hline
    SI-ADV$^w$ &0.0325&\textbf{0.0021}&1.2536&0.0161&\textbf{0.0012}&1.5381 \\
    SI-ADV$^b$ &0.0453&0.0038&1.5702&0.0511&0.0015&1.9875 \\
    Ours &\textbf{0.0273} & 0.0028  &\textbf{1.0126}&\textbf{0.0157}&0.0031&\textbf{1.2848}\\ \hline
    \end{tabular}}}
    \caption{Comparison on the \textbf{ShapeNetPart dataset}.}
    \vspace{-5pt}
    \label{tab:shapenet}
\end{table}

\begin{table}[t!]
\small
    \centering

    \scalebox{0.9}{
    \setlength{\tabcolsep}{1.4mm}{
    \begin{tabular}{c|ccc|ccc}
    \hline
    \multirow{2}*{Method} & \multicolumn{3}{c|}{PointTransformer [B]} & \multicolumn{3}{c}{Point-BERT [C]} \\
    & $D_h$ & $D_c$ & $D_{norm}$ & $D_h$ & $D_c$ & $D_{norm}$ \\ \hline
    SI-ADV$^w$ & 0.0491 & 0.0052  &1.0151&0.0385&0.0028&\textbf{1.2403} \\
    SI-ADV$^b$ &0.0543 & 0.0039  &\textbf{0.8312}&0.0672&\textbf{0.0027}&1.4317\\
    Ours &\textbf{0.0243} & \textbf{0.0035}  &0.8635&\textbf{0.0294}&0.0047&1.2618 \\ \hline
    \end{tabular}}}
    \caption{Comparison on the \textbf{ScanObjectNN dataset}.}
    \vspace{-5pt}
    \label{tab:scanobject}
\end{table}

\noindent \textbf{Experiment of defense method.}
We conduct an experiment on a novel defense method: Lattice Point Classifier (LPC) \cite{li2022robust}. Our 3DHacker achieves a better attack than SI-Adv$^b$ \cite{huang2022shape} as we generate the adversarial sample with high similarity to the original one in both geometric topology and local point distributions.
\begin{table}[t!]
\small
    \centering
    \scalebox{0.9}{
    \begin{tabular}{c|c|cccc}
    \hline
    \multirow{1}*{Model} &\multirow{1}*{Method} & ASR (\%) & $D_h$ & $D_c$ & $D_{norm}$ \\ \hline
    \multirow{2}*{Pointnet} & SI-ADV$^b$ & 82.1 & 0.0458 & \textbf{0.0012} & 2.7804\\
    ~ & ours & \textbf{84.5} & \textbf{0.0146} & 0.0018 & \textbf{1.3519} \\ \hline
    \multirow{2}*{DGCNN} & SI-ADV$^b$ & 65.3 & 0.0421 & \textbf{0.0016} & 1.5804\\
    ~ & ours & \textbf{71.8} & \textbf{0.0213} & 0.0031 & \textbf{1.4652} \\ \hline
    \end{tabular}}
    \caption{Experiment of \textbf{more defense} on Modelnet40.}
    \vspace{-10pt}
    \label{tab:defense}
\end{table}

\end{document}